\documentclass[11pt,a4paper]{article}
\usepackage[hyperref]{ranlp2021}
\usepackage{times}
\usepackage{subcaption}
\usepackage{graphicx}
\usepackage{latexsym}

\usepackage{microtype}

\aclfinalcopy 

\title{BERT Embeddings for Automatic Readability Assessment}

\author{Joseph Marvin Imperial \\
  National University \\
  Manila, Philippines \\
  \texttt{jrimperial@national-u.edu.ph} \\
  }

\date{}

\begin{document}
\maketitle
\begin{abstract}
Automatic readability assessment (ARA) is the task of evaluating the level of ease or difficulty of text documents for a target audience. For researchers, one of the many open problems in the field is to make such models trained for the task show efficacy even for low-resource languages. In this study, we propose an alternative way of utilizing the information-rich embeddings of BERT models with handcrafted linguistic features through a combined method for readability assessment. Results show that the proposed method outperforms classical approaches in readability assessment using English and Filipino datasets—obtaining as high as 12.4\% increase in F1 performance. We also show that the general information encoded in BERT embeddings can be used as a substitute feature set for low-resource languages like Filipino with limited semantic and syntactic NLP tools to explicitly extract feature values for the task.
\end{abstract}

\section{Introduction}
Automatic readability assessment is the task of evaluating the level of ease or difficulty of text documents such as web articles, story and picture books, test materials, and medical prescriptions. Often readability levels can be expressed in many forms: discrete values with grade and age levels such as in the Common European Framework of Reference for Languages (CEFR)\footnote{https://www.cambridgeenglish.org/exams-and-tests/cefr/}, or with continuous values from a given range such as in the famous Lexile Reading Framework\footnote{https://lexile.com/}. In machine learning setting, this task is most often viewed as a classification task where an annotated set of corpora is trained with its corresponding gold-standard labels evaluated by an expert as mostly done in previous works \cite{vajjalatrends,chatzipanagiotidis-etal-2021-broad,weiss-meurers-2018-modeling,xia-etal-2016-text,reynolds-2016-insights,hancke-etal-2012-readability,vajjala-meurers-2012-improving}. Recent works have tried testing unexplored resources by utilizing large pre-trained language models such as Bidirectional Encoder Representations or \textbf{BERT} \cite{devlin-etal-2019-bert} which is based on the attention-driven Transformer architecture by \citet{vaswani-et-al} by (a) directly processing the data to the network \cite{martinc2021supervised,tseng2019innovative} or by (b) using the discrete output of the network via transfer learning \cite{deutsch-etal-2020-linguistic} as an additional feature. For these methods, however, evidence of efficacy are only seen in high-resources readability datasets in English. Thus, we propose an alternative way of incorporating the knowledge of large language models such as BERT by combining its information-rich sentence embeddings as a separate feature set for traditional machine learning algorithms with handcrafted linguistic features. We argue that this method is not only low-resource friendly but also preserves the semantic and syntactic information encoded by the attention heads of BERT since the embeddings itself will be used. We show that such information can act as a substitute for languages with limited tools for explicitly extracting semantic and syntactic features where results describe non-significance in difference of performances between models using semantic and syntactic features versus models using BERT embeddings.

\section{Previous Work}
The first generation of readability formulas and indices date as early as 1920-1940s with the works of \citet{thorndike1921teacher}, \citet{dale1948formula}, and \citet{flesch1948new} primarily using surface-based variables such as raw frequencies and average values of sentences and words per document. The process for using such indices requires manual computation and plugging of values to formulas which can be tedious as the length of a document increases. Likewise, experts argue that considering narrow, surface-based features do not entirely capture the linguistic complexity of a given text \cite{Macahilig2015,collins-thompson-callan-2004-language,SiCallan}. Thus, incorporation of deeper, linguistic variables such as a language's semantics, syntax, morphology, and discourse properties are imperative and worth exploring for the task. To answer this call, the use of handcrafted linguistic features remained the most popular type of input for training readability assessment models through the years. Handcrafted linguistic features are often represented as real-valued numbers serving as potential predictors of the difficulty of reading materials. These features span on a wide range of linguistically motivated factors that base on syntax, semantics, morphology, cohesion, and cognition to name a few. These features also serve as the input in the form of vectors for conventional readability assessment setups using traditional classification-based algorithms. To note, not all linguistic features can be applied or extracted for all languages as some have limited NLP tools suitable for use especially for low-resource languages. Notable works in various languages such as Greek \cite{chatzipanagiotidis-etal-2021-broad}, German \cite{weiss-meurers-2019-analyzing,weiss-meurers-2018-modeling,hancke-etal-2012-readability}, Bangla \cite{sinha-etal-2012-new-readability}, and Filipino \cite{imperial2021application,imperial2020exploring} have used this approach in combination with traditional machine learning algorithms such as Logistic Regression and Support Vector Machines. Likewise, another reason why studies have resorted to the classical approach of model building is that deep neural models are not practical for the task without a large amount of training data.

The advent of large and complex pre-trained language models such as BERT and its variations spawned a handful of studies on how these models fare with the readability assessment tasks. The work of \citet{martinc2021supervised} on the supervised experiment setup explored directly using English benchmark corpus such as Weebit and OneStopEnglish as input for BERT via transfer learning while \citet{deutsch-etal-2020-linguistic} explored using the final discrete output of BERT as a feature for the same datasets. Results from both studies show effectiveness of BERT for English data as direct input while no significant improvement is seen when the discrete output itself is used as a feature. While these results are remarkable, BERT's effectiveness remain a gray area for low-resource languages.

\section{Task Definition}
We define our task at hand as a supervised learning setup. Given a text document $d$ where a feature vector $x = [x_{1},x_{2}\ldots,x_{n}]$ is extracted, a model $M$ is trained using said collection of features $X$ along with the gold label $Y$ or expert-identified readability level. The label is relative in form (discrete or continuous) based on how readability levels are categorized for each corpus.


\begin{table}[!htbp]\small
\centering
\begin{tabular}{|l|c|c|c|}

\hline \bf Data & \bf Doc Count & \bf Sent Count &\bf  Vocab\\  \hline

OSE & 567 & 4,890 & 17,818\\ \hline
CCE & 168 & 20,945 & 78,965\\ \hline
Adarna House & 265 & 10,018 & 16,058\\ \hline

\end{tabular}
\caption{\label{tab:data}
Data distribution for English and Filipino corpus.}
\end{table}

\section{Corpus}
We describe each corpus used in the study below as well as the statistics and breakdown in Table \ref{tab:data}\linebreak

\noindent\textbf{OneStopEnglish.} The OSE corpus is a collection of 567 texts in three different reading levels (beginner, intermediate, and advanced) for adult ESL learners from the MacMillan Education website\footnote{https://www.onestopenglish.com/}. This corpus was first used in the work of \citet{vajjala-lucic-2018-onestopenglish} and has become one of the most-used benchmark datasets for readability assessment and text simplification in English.\linebreak

\noindent\textbf{Common Core Exemplars.} The CCE dataset contains 168 prose texts from the Appendix B of the Common Core State Standards Initiative (CCSS)\footnote{http://www.corestandards.org/assets/Appendix\_B.pdf} for English Language studies and first used by \citet{flor-etal-2013-lexical} for readability assessment. The initiative was a project of the National Governors Association and the Council of Chief State School Officers in USA\footnote{http://www.ccsso.org}. The dataset is divided into three age-range categories: 2-5, 6-7, and 9-12. \linebreak

\noindent\textbf{Adarna House.} The Adarna House corpus is a collection of 265 story books for grades 1-3 from Adarna House Inc.\footnote{https://adarna.com.ph/}, the largest children's literature publisher in the Philippines. This corpus has been used by \citet{imperial2019developing, imperial2020exploring, imperial2021application} for readability assessment in Filipino\footnote{Filipino is considered as a low-resource language \cite{cruz-etal-2020-resource,cruz-etal-2020-localization}.}


\begin{figure*}[!htbp]
    \centering
    \includegraphics[width=12cm]{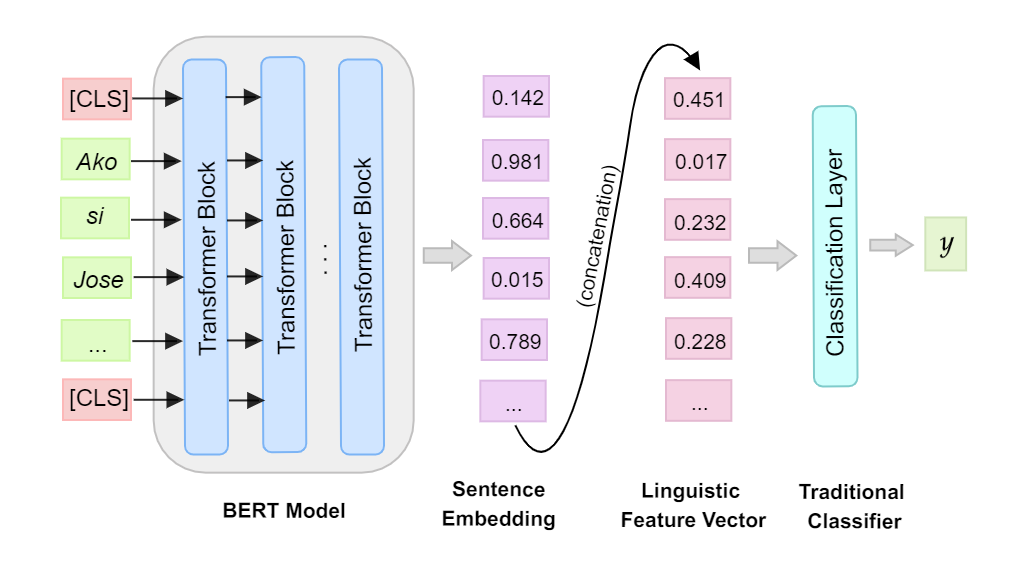}
    \caption{The proposed combined training approach using sentence embeddings from BERT model and extracted handcrafted linguistic feature sets.}
    \label{fig:methodology}
\end{figure*}

\section{BERT Embeddings + Handcrafted Linguistic Features}
BERT's efficacy on a wide range of NLP tasks stems from its implicit capability to encode linguistic knowledge such as hierarchical parse trees \cite{hewitt-manning-2019-structural}, parts of speech and syntactic chunks \cite{liu-etal-2019-linguistic,tenney2019}, semantic roles \cite{ettinger2019} as well as entity types and relations \cite{tenney2019} to name a few. In view with this, we find such amount of knowledge an extremely valuable resource which can potentially improve performances of readability assessment models especially for low-resource languages if used correctly. Thus, to maximize the potential of BERT for low-resource readability assessment, we propose a combined training of its raw embeddings with handcrafted linguistic feature sets through a \textit{concatenation process} and feeding them to traditional machine learning algorithms. The embeddings of BERT generated by the multi-head attention layers are \textit{information-rich}, specifically on semantic and syntactic knowledge \cite{rogers-etal-2020-primer}, due to the nature of its training. We describe our proposed architecture in Figure~\ref{fig:methodology} with a sample Filipino sentence for context.

\section{Experiment Setup}
For the OSE and CCE corpus in English, we extracted over 155 linguistic features covering lexical diversity and density features, syntactic features based on parse trees, morphosyntactic properties of lemmas, and word-level psycholinguistic features. For the Adarna House corpus in Filipino, we extracted over 54 linguistic features covering traditional surface-based features, lexical features based on POS tags, language model features, morphology based on verb inflection, and orthographic features based on syllable pattern. The size of the BERT embeddings for all datasets remain equal with a fixed dimension of $H$ = 768 since the base version of BERT for English \cite{devlin-etal-2019-bert} and Filipino \cite{localization2020cruz,establishing2020cruz,evaluating2019cruz} were used. The embeddings and extracted linguistic feature sets were concatenated, for a total of 923 dimensions for combined features for both English datasets and 823 for the Filipino dataset. Recipes for feature extraction were obtained from the studies of \citet{vajjalaM16,vajjala2014ReadabilityAF} for English and \citet{imperial2020exploring,imperial2021application,imperialASimple2021} for Filipino. We used the \texttt{sentence-transformers} library by \citet{reimers-2019-sentence-bert} with mean pooling option to extract BERT embedding representations for the readability corpora\footnote{We release the script for extracting BERT embeddings at \url{https://github.com/imperialite/BERT-Embeddings-For-ARA}}.

For the traditional machine learning algorithms, we used three of the commonly utilized in previous works: Logistic Regression, Support Vector Machines, and Random Forest. Models for each dataset were trained on a 5-fold cross validation procedure. We used weighted F1 as the overall metric for performance evaluation.

\section{Results}

\subsection{Ablation}
We compared performances of models on three different setups, (a) linguistic features only, (b) BERT sentence embeddings only, and (c) combined training of the two feature embeddings to gauge the efficacy of the proposed framework.


\begin{table*}[!htbp]

    \begin{subtable}[h]{\textwidth}
    \centering
    \begin{tabular}{|l|c|c|c|}
    
    \hline \bf Method & \bf OSE & \bf CCE &\bf  Adarna \\  \hline
    
    Linguistic Features                 & 0.676     & 0.774     & 0.389 \\ \hline
    BERT Embeddings                     & 0.620     & 0.747     & 0.505 \\ \hline
    \bf Combined Features (Ling + BERT)    & \bf 0.732     & \bf 0.778     & \bf 0.554 \\ \hline
    
    \end{tabular}
    \vspace{0.2 cm}
    \subcaption{Logistic Regression}
    \label{tab:logreg}
    \end{subtable}
    
    \vspace{0.5 cm}
    
    \begin{subtable}[h]{\textwidth}
    \centering
    \begin{tabular}{|l|c|c|c|}
    
    \hline \bf Method & \bf OSE & \bf CCE &\bf  Adarna \\  \hline
    
    Linguistic Features                 & 0.691     & 0.732     & 0.414 \\ \hline
    BERT Embeddings                     & 0.611     & 0.826     & 0.487 \\ \hline
    \bf Combined Features (Ling + BERT)    & \bf 0.704     & \bf 0.893     & \bf 0.571 \\ \hline
    
    \end{tabular}
    \vspace{0.2 cm}
    \subcaption{Support Vector Machines}
    \label{tab:svm}
    \end{subtable}

    \vspace{0.5 cm}
    
    \begin{subtable}[h]{\textwidth}
    \centering
    \begin{tabular}{|l|c|c|c|}
    
    \hline \bf Method & \bf OSE & \bf CCE &\bf  Adarna \\  \hline
    
    Linguistic Features                 & 0.683     & 0.842     & 0.423 \\ \hline
    BERT Embeddings                     & 0.439     & 0.770     & \bf 0.504 \\ \hline
    \bf Combined Features (Ling + BERT)    & \bf 0.690     & \bf 0.861     & 0.467 \\ \hline
    
    \end{tabular}
    \vspace{0.2 cm}
    \subcaption{Random Forest}
    \label{tab:rf}
    \end{subtable}

\caption{F1 performance via training with (a) Logistic Regression, (b) Support Vector Machines, and (c) Random Forest using handcrafted linguistic features, BERT sentence embeddings, and combined training of both.}
 \label{tab:experiments}
\end{table*}

As described in Table~\ref{tab:experiments}, generally speaking, models trained using the proposed combined training of handcrafted linguistic feature sets with contexual BERT embeddings outperform both performances of only using each exclusively on English and Filipino datasets. On average, we note an increase of performance of 2.63\% for OSE, 6.23\% for CCE, and 12.4\% in weighted F1 score for Adarna House across all algorithms. From this, we infer that extracting and incorporating the information-rich embeddings of any readability dataset using BERT to commonly-used linguistic feature sets can substantially improve model performance. 

Interestingly, there are a few notable cases reported in Table~\ref{tab:experiments} where BERT embeddings alone outperformed the traditional method of using handcrafted linguistic feature sets as primary input. These cases are evident in the all models utilizing the Adarna House dataset in Filipino with an average increase of 9.5\% weighted F1 scores. From this, we infer that the general semantic and syntactic knowledge implicitly encoded in BERT embeddings as detailed in probing tasks from previous works \cite{rogers-etal-2020-primer,hewitt-manning-2019-structural,liu-etal-2019-linguistic,tenney2019} may be significantly more informative than the traditional handcrafted linguistic features for discriminating reading difficulty. Consequently, this poses as a probable and alternative solution for low-resource languages with little to no NLP tools such as a good part-of-speech tagger, stemmer, syntactic parse tree extractor, and morphological analyzer to name a few for manually extracting linguistic information from documents. Since BERT models are trained in an self-supervised manner, the overhead of developing these tools from scratch can be disregarded, at least for readability assessment. We discuss further experiments on this inference in the next section.


\begin{table*}[!htbp]
    \centering
    \begin{tabular}{|l|c|c|c|}
    
    \hline \bf Model w/ Removed Features & \bf OSE & \bf CCE &\bf  Adarna \\  \hline
    
    Logistic Regression     & \bf 0.744     & 0.865     & 0.492 \\ \hline
    Support Vector Machines                     & 0.615     & 0.869     & 0.507 \\ \hline
    Random Forest           & 0.669     & 0.791     & 0.431 \\ \hline
    \hline
    Full Model (Ling + BERT) & 0.732     & \bf 0.893     & \bf 0.571 \\ \hline
    
    \end{tabular}

\caption{Performances of models via F1 score after retraining with semantic and syntactic handcrafed linguistic features removed to test if information-rich BERT embeddings can act as substitution for such features. Best performing model utilizing combined features from Table~\ref{tab:experiments} appended for comparison.}
\label{tab:substitution}
\end{table*}

\begin{figure*}[!htbp]
    \centering
    
    \begin{subfigure}[b]{\linewidth}
        \includegraphics[width=.29\textwidth,trim={0.4cm 0.4cm 0.4cm 0.4cm},clip]{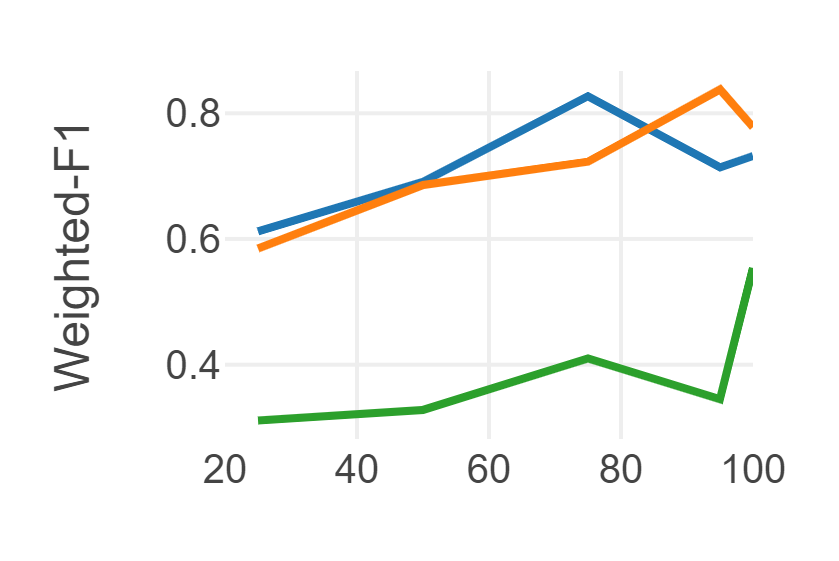}\hfill
        \includegraphics[width=.32\textwidth,trim={0.4cm 0.4cm 0.4cm 0.4cm},clip]{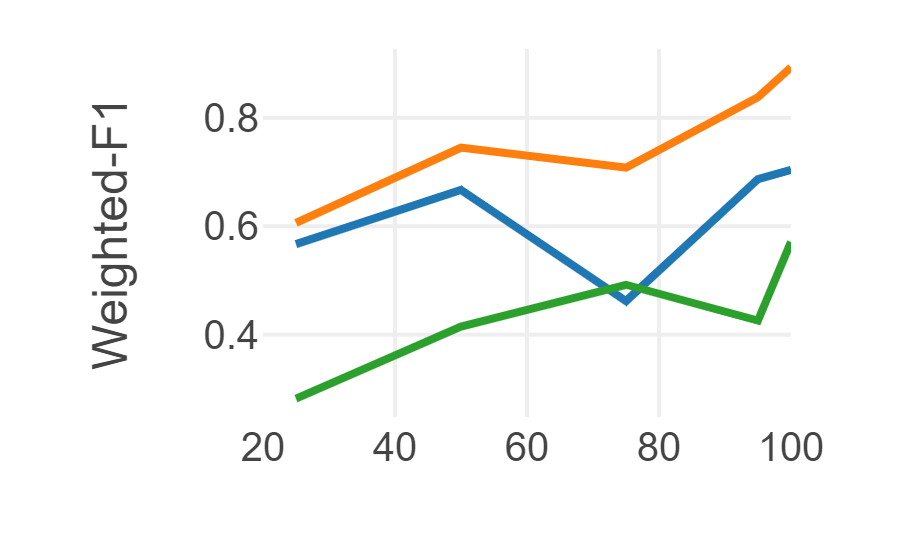}\hfill
        \includegraphics[width=.38\textwidth,trim={0.3cm 0.3cm 0.3cm 0.3cm},clip]{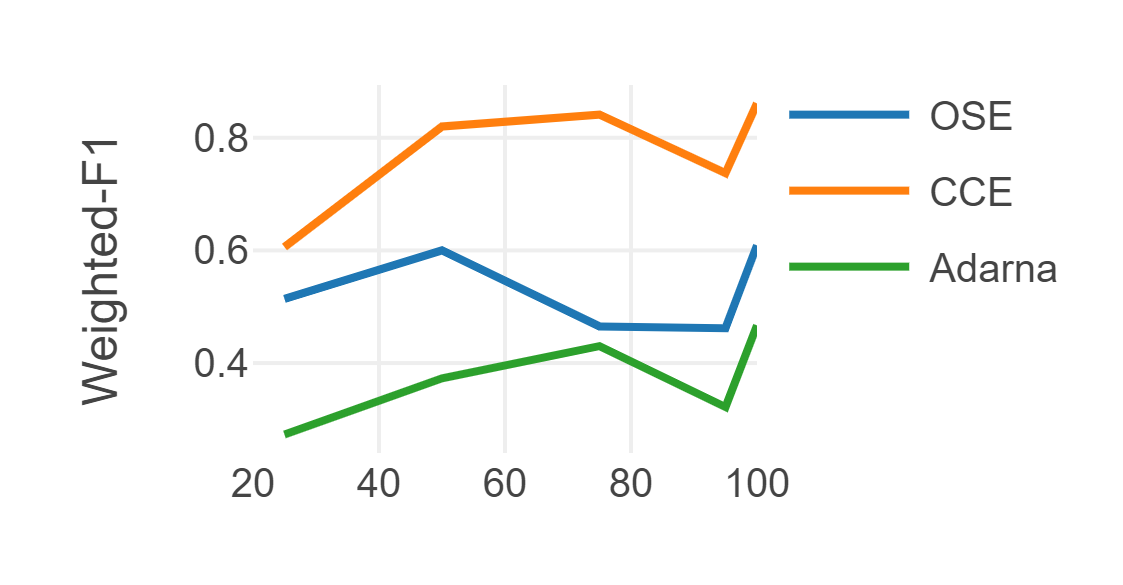}
    \end{subfigure}

    \caption{Decomposing large feature sets on a 25\%, 50\%, 75\%, 95\%, and 100\% (full) variance percentages using PCA for Logistic Regression, Support Vector Machines, and Random Forest (left to right).}
    \label{fig:pca}
\end{figure*}

\subsection{Substituting Semantic and Syntactic Features for BERT Embeddings}
To empirically test if BERT embeddings can act as substitute for semantic and syntactic linguistic features for readability assessment, we removed features from the three datasets that assume semantic and syntactic knowledge. For OSE and CCE, we removed 56 features covering part-of-speech densities, lexical richness, type-token densities, and general parse-tree based features. For Adarna, we removed 22 features covering part-of-speech densities, type-token densities, and verb inflection densities. There are no parse-tree based features for Adarna House as there are currently no NLP tools for extracting such feature set for Filipino. The rest of the linguistic features from the datasets denoting other aspects of reading difficulty measurement such as frequency-based features and syllable patterns remain unchanged. Models were retrained using the three selected machine learning algorithms for comparison.

Results of substitution experiments can be found in Table~\ref{tab:substitution}. Generally speaking, it is evident that models trained using the combined method still outperforms models using the reduced feature set on the account of CCE and Adarna data. However, we note the 1.2\% increase in F1 score on the OSE data. Stemming from this observation, we also note small differences in performances of using the combined features against decreased features. In the CCE corpus, the highest performing model using decreased features obtained 86.9\% F1 score which is less 2.4\% than using the model with combined features. For the Adarna data, the difference is 6.4\%.

To identify if such difference is significant, we used a two-tailed test of difference via Mann-Whitney U using the performance scores of models with combined features and models with decreased features for all datasets. We arrived at a $p$-value of 0.522 ($p>$0.5), meaning that the difference of the scores between two groups is not significant\footnote{The distribution of the two groups are of equal variances with $p$-value of 0.619.}. Thus, we conclude that BERT embeddings can be fully used as a substitute for semantic and syntactic features if such information cannot be explicitly extracted from readability data due to the lack of NLP tools and low-resourceness of other languages. To add, since BERT models are trained in an self-supervised manner and there are over 3,000 pretrained models from online repositories\footnote{https://huggingface.co/models}, these resources and the proposed combined training method as a viable option.

\subsection{Feature Decomposition for Performance Boost}
In extending the effort to improve the performance of BERT-enriched readability assessment models and reduce feature size or dimensionality, we resorted to the use of feature decomposition to the large feature vector sizes (BERT + linguistic features) via Principal Components Analysis (PCA). PCA works by projecting the overall feature set (often large) to a lower dimensional property while preserving quality and information of features \cite{Hotelling1933AnalysisOA}. We experimented on differing values of variance percentages: 25, 50, 75, 95, and 100 (full, no feature removed). Results of feature decomposition via PCA for each machine learning model are described in Figure~\ref{fig:pca}. 

For SVM and Random Forest, all datasets have the highest performances if all features are retained (100 variance percentage). While for Logistic Regression, 75 variance percentage obtained the highest performance with 82.7 F1 score for OSE, 95 variance percentage obtained the highest performance with 83.8\% F1 score for CCE, and 100\% or full features for Adarna. Thus, we infer that there is no need to perform feature decomposition to find the principal components as the highest-performing models for OSE, CCE, and Adarna use 100\% of the combined feature set (BERT + linguistic features).

\section{Conclusion}
In this study, we proposed an alternative way of combining information-rich BERT embeddings with handcrafted linguistic features for the readability assessment task. Results from our experiments showed that the method outperforms classical, vanilla approaches in readability assessment using English (OSE and CCE) and Filipino (Adarna) datasets in various machine learning algorithms such as Logistic Regression, Support Vector Machines, and Random Forest. We also demonstrated that the knowledge implicitly encoded in BERT embeddings (semantic and syntactic information) can be used as a full substitute feature set for low-resource languages like Filipino with limited NLP tools to explicitly extract feature values for the task. We are looking forward to the application of our proposed method to other languages struggling with the extraction deep linguistic features to trained readability assessment models. Future directions of the study include deeper exploration of BERT such as isolating extracted embeddings for each of the twelve attention layers. 

\section{Acknowledgments}
We would like to thank Dr. Ani Almario from Adarna House, Dr. Sowmya Vajjala from the National Research Council of Canada, and Dr. Michael Flor from ETS for providing the Adarna, OSE, and CCE datasets respectively.

\section{Ethical Considerations}
We report that there are no major ethical concerns in the study as it involves no human subjects nor discriminate any identifiable group of people. As for the dataset, for Adarna House, permission was obtained from the publishing house while for the OSE and CCE datasets, it remains open-sourced. As for energy consumption, the study only uses pre-trained BERT models and the authors did not actually perform the pre-training phase itself.

\bibliographystyle{acl_natbib}
\bibliography{anthology,references}


\end{document}